\pdfoutput=1

\documentclass[11pt]{article}

\usepackage[]{EMNLP2023}

\usepackage{times}
\usepackage{latexsym}
\usepackage{amsmath}

\usepackage[T1]{fontenc}

\usepackage[utf8]{inputenc}

\usepackage{microtype}

\usepackage{inconsolata}

\usepackage{mathtools}
\usepackage{booktabs}
\usepackage{multirow}
\usepackage{enumitem}
\usepackage{makecell} 
\usepackage{tcolorbox}
\usepackage{hhline}
\usepackage{graphicx}
\usepackage[export]{adjustbox}
\usepackage{arydshln}

\newcommand{\method}{\textsc{case}}
\newcommand{\methodnoc}{\textsc{cas}}

\title{\method{}: Commonsense-Augmented Score with an Expanded Answer Space}

\author{Wenkai Chen \and Sahithya Ravi \and Vered Shwartz \\
University of British Columbia\\
Vector Institute for AI\\
{\tt \{wkchen, sahiravi,  vshwartz\}@cs.ubc.ca
}}

\begin{document}
\maketitle
\begin{abstract}

LLMs have demonstrated impressive zero-shot performance on NLP tasks thanks to the knowledge they acquired in their training. In multiple-choice QA tasks, the LM probabilities are used as an imperfect measure of the plausibility of each answer choice. One of the major limitations of the basic score is that it treats all words as equally important. We propose CASE, a Commonsense-Augmented Score with an Expanded Answer Space. CASE addresses this limitation by assigning importance weights for individual words based on their semantic relations to other words in the input. The dynamic weighting approach outperforms basic LM scores, not only because it reduces noise from unimportant words, but also because it informs the model of implicit commonsense knowledge that may be useful for answering the question. We then also follow prior work in expanding the answer space by generating lexically-divergent answers that are conceptually-similar to the choices. When combined with answer space expansion, our method outperforms strong baselines on 5 commonsense benchmarks. We further show these two approaches are complementary and may be especially beneficial when using smaller LMs.
\end{abstract}

    \section{Introduction}
\label{sec:intro}
\definecolor{teagreen}{rgb}{0.82, 0.94, 0.75}
\definecolor{tealgreen}{rgb}{0.0, 0.51, 0.5}

\begin{figure}[!ht]
    \centering
    \small
    \tcbox[on line,boxsep=0pt,left=1pt,right=1pt,top=1pt,bottom=1pt,colback=teagreen,colframe=tealgreen,toprule=0.5pt,bottomrule=0.5pt,leftrule=0.5pt,rightrule=0.5pt]{
    \begin{tabular}{l}
         The woman hired a lawyer because \underline{~~~~~}\\
        \textbf{A. she decided to sue her employer.}\\
        B. she decided to run for office.\\
        \cdashline{1-1}
        C. she wanted to sue her former employer.\\
    \end{tabular}
    }
    \includegraphics[width=0.45\textwidth,trim={0 0 0 6.8cm},clip]{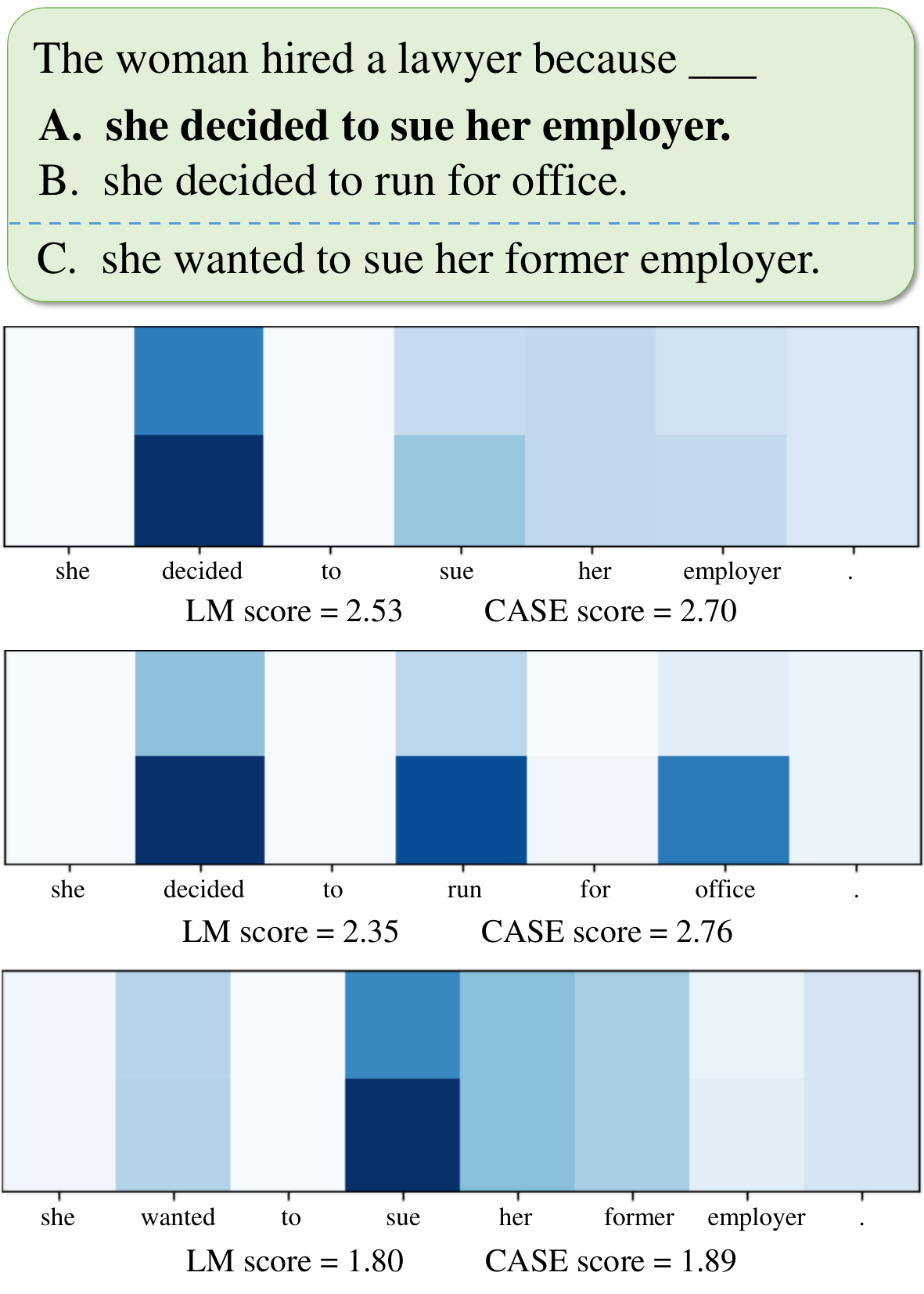}
    \caption{An example from COPA. A and B are the original options, while option C was generated by GPT-2 as part of the answer space expansion step. The top line in each heatmap represent the LM (cross-entropy) score and the bottom line represents our \method{} score. Higher scores and blue blocks correspond to lower plausibility. CASE correctly predicts option A (and option C which is an expansion of A) as more plausible than option B, while the LM-score incorrectly predicts option B.}
    \label{fig:overview}
\end{figure}


Large language models (LLMs) have demonstrated strong few-shot and zero-shot performance across various NLP tasks, with the larger models often matching earlier fine-tuned approaches that relied on task-specific labeled data \cite{gpt2, gpt3, touvron2023llama}. We focus on the zero-shot setup, which assumes that the knowledge needed to perform a specific task is already present in the LLM \cite{petroni-etal-2019-language, zhou2020evaluating, Saha_Joty_Hoi_2022}. Zero-shot learning has been employed for tasks such as translating between unseen language pairs \cite{zhang-etal-2020-improving}, summarization \cite{gpt3},  commonsense reasoning \cite{shwartz-etal-2020-unsupervised,klein-nabi-2021-towards,liu-etal-2022-generated, fang-etal-2022-leveraging}, and more.

In multiple-choice question answering (MCQA) tasks, zero-shot methods typically rely on the language model (LM) probabilities as a proxy for plausibility, predicting the answer choice with the highest probability conditioned on the question. LM score is a na\"{i}ve proxy for plausibility, since it confounds factors such as length, unigram frequency, and more \cite{holtzman-etal-2021-surface,niu-etal-2021-semantic}. Indeed, in Figure~\ref{fig:overview}, a GPT-2 based LM score incorrectly predicts that the woman hired a lawyer because she decided to run for office, rather than because she decided to sue her employer. 

In this paper, we propose to address one of the major limitations of the LM score. By summing or averaging the token-level probabilities, the LM score treats all tokens as equally important. A person reading this question would likely pay attention to option A because the word ``sue'' is highly relevant in the context of a lawyer. This signal might be weaker in a basic LM score where the word ``sue'' is conditioned on each other token in the question and previous tokens in the answer. Furthermore, the LM might miss non-trivial connections between related words. 

To address this challenge, we propose \method{}: a \textbf{C}ommonsense-\textbf{A}ugmented \textbf{S}core with an \textbf{E}xpanded Answer Space. \method{} is a post-hoc dynamic weight scoring algorithm that prioritizes important words in the sentence. The importance of each individual word is determined based on its relationship with other words in ConceptNet \cite{speer2017conceptnet}. For example, ConceptNet provides the information that ``sue requires having a lawyer''. We use the word-level importance scores to re-weigh the LM probability scores. Indeed, in the second line of option A in Figure~\ref{fig:overview}, the importance of the word ``sue'' increases the score of the entire sentence, leading to correctly predicting A as the correct answer. 

We further adopt the strategy suggested by \newcite{niu-etal-2021-semantic} to expand the answer space by using a LM to generate additional answers and then mapping semantically-similar generated answers into the original space. This mitigates the LM score's sensitivity to infrequent words. Figure~\ref{fig:overview} demonstrates that a generated option C, ``she wanted to sue her former employer'', which is conceptually similar to A, further yields a higher probability score with our method.

We tested \method{} on 5 popular commonsense MCQA datasets. \method{} outperformed the broad range of strong baselines that we compared with, confirming that it is an effective method for zero-shot MCQA. We further study the impact of different model sizes, answer candidates of varying qualities, and different weight assignment strategies on the performance.\footnote{Our code is available at \href{https://github.com/WK-Chen/Commonsense-Augmented-Score-with-an-Expanded-Answer-Space}{Github.}}

\section{Background}
\label{sec:bg}


\subsection{Plausibility Scoring}
\label{sec:bg_plausibility_score}

Although the plausibility score of a sentence can be easily calculated by accumulating the probability assigned by the LM for each token, this approach suffers from various statistical biases such as sensitivity to the number of tokens, subword tokenization, and word frequency \cite{abdou-etal-2020-sensitivity,holtzman-etal-2021-surface}. To address these biases, several improvements have been proposed. With respect to the length bias, prior work normalized the score by length \cite{mao-etal-2019-improving,brown2020language}, or focused on the conditional probabilities of the question, which unlike the answer choices has a fixed length  \cite{trinh2018simple, tamborrino-etal-2020-pre}. To factor out word frequency, \newcite{holtzman-etal-2021-surface} proposed Domain Conditional Pointwise Mutual Information (DCPMI), which normalizes the conditional probability of the answer given the question by the prior probability of the answer. This is computed as the conditional probability of the answer given a domain-specific prefix such as ``The sentiment of the movie is'' for sentiment analysis or ``The answer is'' for general QA tasks. SEQA \cite{niu-etal-2021-semantic} mitigates the sensitivity to word choice by generating answers using GPT-2, and selecting the answer choice most similar to the generated answers.

Existing methods solely focus on the relationship between words in the choices and words in the question, ignoring the importance of each word for the decision. In this paper, we propose a new token-level weighting method to consider the importance of different words within the sentence based on their relationship to other words. 

\begin{figure*}[t]
    \centering\includegraphics[width=2.0\columnwidth]{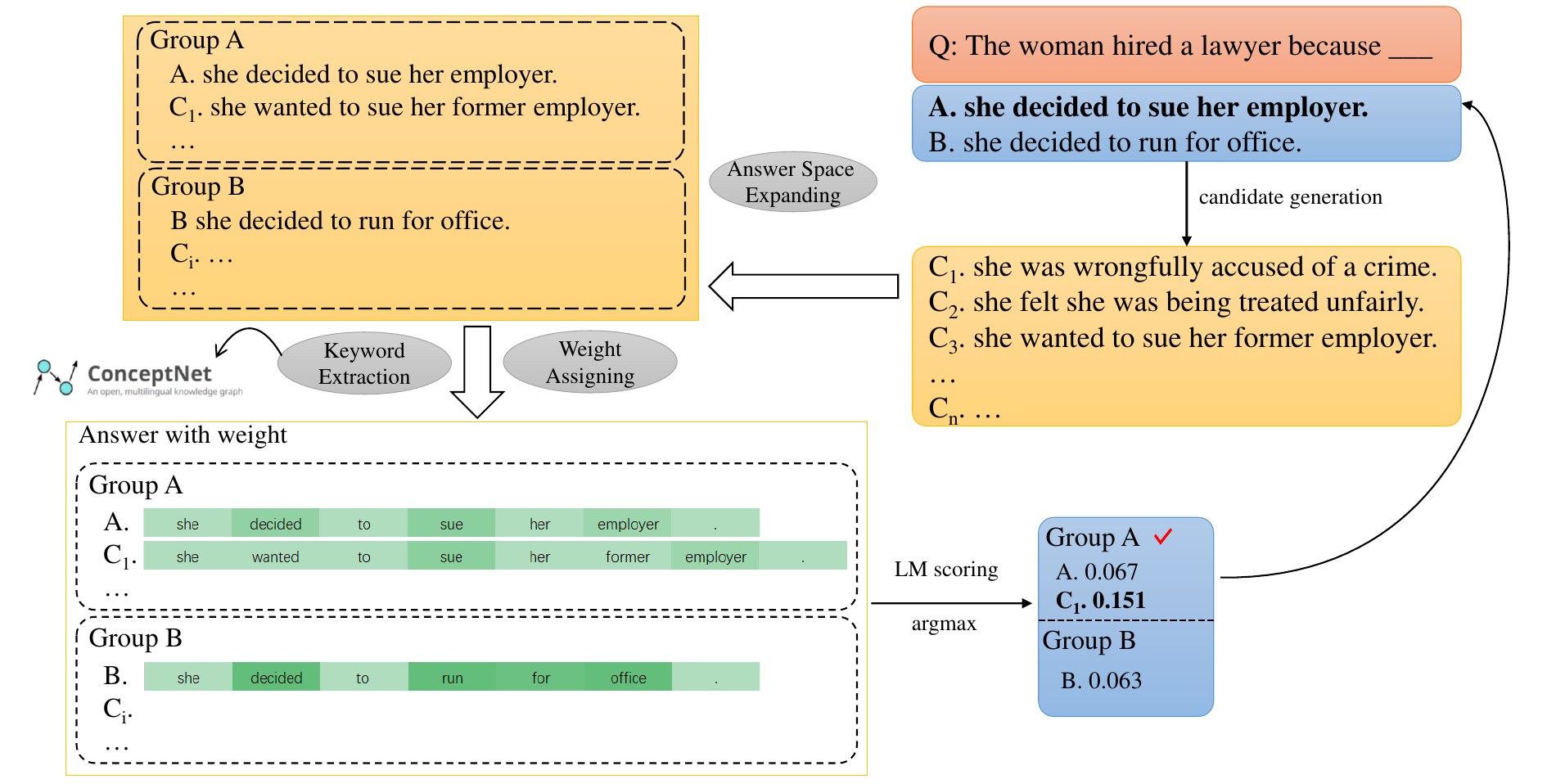}
    \caption{Overview of \method{}, illustrated with an example from the COPA dataset.
    Groups A and B correspond to original choices A and B and  any generated answers mapped to them (\S\ref{sec:method_generation}). Each word in each answer is scored based on its ConceptNet relationships to other words in the instance (\S\ref{sec:method_weight}). The score for each answer is based on the word probabilities (\S\ref{sec:method_basic}), weighted by the word-level scores. Finally, \method{} predicts the answer choice with the highest scoring answer in its group.}
    \label{fig:framework}
\end{figure*}

\subsection{Knowledge-Enhanced Models}
\label{sec:bg:external_knowledge}

Zero-shot LM-based scoring methods implicitly reason about which answer is more likely based on the token-level probabilities. However, many tasks require multiple steps of reasoning to reach the correct answer \cite[e.g.,][]{mihaylov-etal-2018-suit, yang-etal-2018-hotpotqa, khot2020qasc}. A common approach is to retrieve  relevant commonsense knowledge from knowledge bases (KBs) such as ConceptNet \cite{speer2017conceptnet} and ATOMIC \cite{sap2019atomic,Hwang2021COMETATOMIC2O}, in order to enhance the neural model and explicate the reasoning steps 
\cite[e.g.,][]{bauer-etal-2018-commonsense,xia2019incorporating,lin-etal-2019-kagnet,Guan_Wang_Huang_2019, chen-etal-2020-improving, huang-etal-2021-improving}. More recent work used the COMET model \cite{bosselut-etal-2019-comet,Hwang2021COMETATOMIC2O}, which is a LM fine-tuned on the aforementioned KBs, to enhance models with high-coverage contextualized commonsense inferences \cite[e.g.,][]{majumder-etal-2020-like,bosselut2021dynamic,kim-etal-2022-mind,chakrabarty-etal-2022-rocket,Ravi_2023_WACV}. 

An alternative recent approach which doesn't rely on external KBs prompts a LM to generate additional knowledge which is then incorporated back into the LM to make the prediction.  \newcite{shwartz-etal-2020-unsupervised} and later \newcite{liu-etal-2022-generated} used a LM to generate questions and answers about an MCQA instance. The answers to the questions are then incorporated into the LM-based scoring model as additional knowledge. \newcite{wei2022chain} proposed the popular chain-of-thought (COT) prompting approach in which the LM is taught through examples to generate multiple steps of reasoning followed by the answer to the question. In the zero-shot version, the LM is instructed to ``think step-by-step''. Finally, following concerns about the faithfulness of CoT inferences, \newcite{creswell2022selection} proposed to iteratively  select parts of the inputs and draw inferences on them. 

\section{Method}
\label{sec:method}
We propose \method{}, a \textbf{C}ommonsense-\textbf{A}ugmented \textbf{S}coring method with an \textbf{E}xpanded Answer Space. \method{} can be used for zero-shot MCQA tasks. It is based on LM score (Section~\ref{sec:method_basic}). However, rather than treating all words in the context and answers as equally important, we propose a weighted score where the conditional probability is weighed by the importance of a word. The weights are determined using a commonsense KB in order to provide information that humans might implicitly be reasoning about when answering such questions (Section~\ref{sec:method_weight}). Following \newcite{niu-etal-2021-semantic}, we expand the set of answer candidates by generating free-text answers, to increase the scorer's robustness to lexical variability (Section~\ref{sec:method_generation}). An overview of the method is shown in Figure~\ref{fig:framework}. 

\subsection{Basic Scoring Method}
\label{sec:method_basic}

The basic scoring method directly uses the LM score, which is calculated by accumulating the conditional probabilities assigned by the LM for each token given the prefix. Given a question $Q = q_1 ... q_{n_Q}$ and an answer choice $A_i = a_{i,1} ... a_{i,{n_{A_i}}}$, we convert $Q$ into a declarative statement $s$ (see Appendix~\ref{sec:appendix_declarative}), and define the LM score of answer choice $A_i$ as follows:
\begin{equation}
\small
\begin{split}
   P_{A_i} &= P(A_i|s) \\ &= \frac{1}{n_s + n_{A_i}} \cdot \prod_{j=1}^{n_{A_i}} P(a_{i,j}|s, a_{i,1}, \cdots , a_{i,{j-1}})
\end{split}
\label{eq:basic}
\end{equation}
\noindent where $n_s$ is the number of tokens in $s$. 


Finally, we can determine the most plausible choice $\hat{A}$ among the answer choices based on their corresponding scores:
\begin{equation}
    \hat{A} = \mathop{\arg\max}\limits_{i} P_{A_{i}}
    \label{eq:argmax1}
\end{equation}

\subsection{Commonsense Augmented Scoring}
\label{sec:method_weight}

The importance of individual words in the question and their contribution to choosing the correct answer varies greatly. Take for example the instance in Figure~\ref{fig:overview}, taken from the COPA dataset \cite{gordon-etal-2012-semeval}. Determining the cause of the event ``The woman hired a lawyer'' involves reasoning about the circumstances in which one might hire a lawyer, such as if they are suing someone. In this case, the keywords ``lawyer'' from the context and ``sue'' from the answer choice, and the semantic relation between them (i.e., suing someone requires a lawyer), supports the correct prediction. To that end, \method{} first identifies important keywords from the question and answer choices (Section~\ref{sec:method_grouping}). Each keyword is assigned an importance score, and the conditional probability $P_A$ is updated 
by considering the importance of each token in the answer choice (Sec~\ref{sec:method_weight_assign}).


\subsubsection{Keywords Extraction}
\label{sec:method_grouping}

Given a question $Q$ and an answer choice $A$, we use YAKE \cite{campos2018yake}, an unsupervised automatic keyword extraction method, to extract a set of keywords $\text{Key}_Q \subset Q$ and $\text{Key}_A \subset A$. In particular, we are interested in finding the keywords from each answer choice that are important in the context of the question $Q$, which we denote $\text{Key}_{A|Q} \subset \text{Key}_A$. To that end, we use ConceptNet \citep{speer2017conceptnet}, a commonsense knowledge base, to find paths between terms in $\text{Key}_Q$ and $\text{Key}_A$, and include in $\text{Key}_{A|Q}$ keywords from the answer choice that are connected in ConceptNet to keywords from the question:
\begin{equation}
\small
\text{Key}_{A|Q} = 
    \left\{
        a \in \text{Key}_A 
        \left| 
        \begin{aligned}
         & \exists q \in \text{Key}_Q~\wedge \\ 
         & \exists p = a \leadsto q \in \text{CN}~\wedge \\ 
         & |p| \le k \\
         \end{aligned} 
        \right.
    \right\} 
\end{equation}


\noindent where $p$ denotes a path in ConceptNet (CN) with up to $k$ edges. 

\subsubsection{Weight Assigning}
\label{sec:method_weight_assign}

We assign a weight to each token $a \in \text{Key}_{A|Q}$ based on the strength of its connection to keywords in $\text{Key}_Q$. 
To that end, we look at all the ConceptNet paths that connect $a$ with keywords in $\text{Key}_Q$, which we denote $\text{Paths}_{a \leadsto}$. We convert the path to a set of sentences by expressing each edge as a natural language sentence, based on relation templates (see Appendix~\ref{sec:appendix:path_sents}). For example, the path $\texttt{sue} \xrightarrow[]{\text{related to}} \texttt{law} \xleftarrow[]{\text{in context of}} \texttt{lawyer}$ is expressed as $S_1$ = ``sue is related to law'' and $S_2$ = ``lawyer is a word used in the context of law''. 
We use the LM to score a single path $P_{a \leadsto q}$ as follows. First, the score $S(E_i)$ of edge $E_i = (x_i, R_i, y_i)$ is calculated as the conditional probability of generating the second node $y_i$ following the textual template of relation $R_i$, to which we assign the first node $x_i$, such as P(law|sue is related to). We use the chain rule for conditional probability to compute the score of the entire path:
\begin{equation}
\small
S(P_{a \leadsto q}) = \frac{1}{|P_{a \leadsto q}|+1} \left(\sum_1^{|P_{a \leadsto q}|} \operatorname{log} S(E_i) + \operatorname{log} S(E') \right)
\end{equation}
\noindent where $E'$ is an artificial summary edge from $x_1$ to $y_{P_{a \leadsto q}}$ with the ``is related to'' relation, such as ``sue is related to lawyer''. 

To get an aggregated score for a token $a$, we sum the scores of all paths in $\text{Paths}_{a \leadsto}$:
\begin{equation}
    S_{\text{Paths}_{a \leadsto}}=\sum_{P_{a \leadsto q} \in \text{Paths}_{a \leadsto}} S(P_{a \leadsto q})
\end{equation}
Finally, the weight for each token $a_{i,j}$ in $A_i$ is computed as follows.
\begin{equation}
\small
    W_{a_{i,j}} =
    \begin{cases}
    1+ \lambda S_{\text{Paths}_{a_{i,j} \leadsto}}, & \text{if } a_{i,j} \in \text{Key}_{A_i|Q} \\
    1, & \text{if } a_{i,j} \notin \text{Key}_{A_i|Q}
    \end{cases}
\end{equation}
\noindent where $\lambda$ is a hyperparameter (\S\ref{sec:exp:setup}). 

With the weights for each token, we can now update the LM score defined in Equation~\ref{eq:basic} to a weight-based plausibility score as follows:
\begin{equation}
    P_{A_i} = \prod_{j=1}^{n} W_{a_{i,j}} \cdot P(a_{i,j}|s, a_{i,1}, \cdots , a_{i,j-1})
\label{eq:weighted_score}
\end{equation}

\subsection{Expanded Answer Space}
\label{sec:method_generation}

The final addition to our model aims at reducing the LM sensitivity to the phrasing of the correct answer. For example, an infrequent word in the correct answer choice can reduce the overall probability of the choice and make the LM predict another option as more plausible \cite{holtzman-etal-2021-surface}. To mitigate this issue, we follow \newcite{niu-etal-2021-semantic} and expand the set of answer candidates by using a causal LM to generate open ended answers $A^* = \{A^{*}_1, ..., A^{*}_{n_{A^*}}\}$. The idea is to allow the model to consider various phrasings of the same conceptual answer. For example, in Figure~\ref{fig:framework},  the generated answer $C_1$ is a paraphrase of answer choice $A$. 


We treat the generated answer choices $A^*$ the same as the original answer choices $A$ and compute the score for each answer $A^{*}_i \in A^*$ using Equation~\ref{eq:weighted_score}.  To map the answer choices back into the original answer space $A$, we attempt to match each $A^{*}_i \in A^*$ to $A_i \in A$ based on two criteria: sentence similarity and keyword connections. 

\paragraph{Sentence Similarity.} We use the Sentence-Transformer package \cite{reimers-gurevych-2019-sentence} to represent the answers, and compute the cosine similarity between the representations of each generated answer in $A^{*}$ and original answer in $A$. The similarity score between the sentence pair should be above $s_{sim}$. 

\paragraph{Keyword Connections.} We calculate the connection score between the keywords in each generated answer in $A^*$ and each original answer in $A$ using the method introduced in Sec~\ref{sec:method_weight_assign}. We require the connection score to be greater than 0.

A candidate can only be assigned to a group if it meets both thresholds, and we discard generated answers that are not mapped into answer choices in $A$. 
Once we mapped generated answers to original answers, the final prediction of the model modifies Equation~\ref{eq:argmax1} to select the highest scores of all answers within the same cluster:
\begin{equation}
    \hat{A} = \mathop{\arg\max}\limits_{i} \mathop{\arg\max}\limits_{j} {P_{A_{i,j}}}
\end{equation}
where $A_{i,j}$ is the $j$th answer in cluster $A_i$.

\section{Experimental Setup}
\label{sec:exp_setup}
\subsection{Datasets}
\label{sec:exp_datasets}

We evaluated our method on five multiple-choice commonsense question answering datasets described below.

\paragraph{COPA.} The goal in  the \textbf{C}hoice \textbf{o}f \textbf{P}lausible \textbf{A}lternatives dataset \cite[COPA;][]{roemmele2011choice} is, given a premise event, to choose the more plausible cause or effect among two alternatives.

\paragraph{SCT.} The \textbf{S}tory \textbf{C}loze \textbf{T}est dataset \cite[SCT;][]{mostafazadeh-etal-2016-corpus} is a collection of four-sentence stories with two possible endings. The goal is to predict which ending is more plausible following the beginning of the story.

\paragraph{SocialIQA.} The \textbf{Social} \textbf{I}nteraction \textbf{Q}uestion \textbf{A}nswering \cite[SocialIQA;][]{sap-etal-2019-social} dataset tests models on their understanding of social situations and human behavior. Each question presents a hypothetical scenario followed by a question and 3 answer choices.

\paragraph{ARC.} The \textbf{A}I2 \textbf{R}easoning \textbf{C}hallenge \cite[ARC;][]{clark2018think} consists of 7,787 science exam questions drawn from a variety of sources. The questions are divided into Easy (ARC-E) and Challenging (ARC-C) sets.

\paragraph{OBQA.} The \textbf{O}pen\textbf{B}ook\textbf{QA} \cite[OBQA;][]{mihaylov-etal-2018-suit} dataset contains questions that require multi-step reasoning, use of commonsense knowledge, and rich text comprehension. The dataset has roughly 6,000 questions.

Since the test set of SCT and SocialIQA are not publicly-available, we report the accuracy on the development set for all datasets.

\subsection{Baselines}
\label{sec:exp:baselines}

We compare our proposed method with the basic LM-based scoring method described in Section~\ref{sec:method_basic}, as well as more advanced  LM-based scoring methods described below. 

\paragraph{Self-talk} \cite{shwartz-etal-2020-unsupervised} consists of two causal LMs. The knowledge generator LM generates clarification questions conditioned on the context and pre-defined prefixes, and their corresponding answers. The scoring LM computes the probability of each answer choice conditioned on the context and question as well as the additionally generated knowledge.\footnote{We don't compare with follow-up work by \newcite{liu-etal-2022-generated} since they targeted a different set of tasks.} 

\paragraph{DC-PMI} \cite{holtzman-etal-2021-surface} aims to eliminate the effect of the number of synonyms and the word frequency on the LM score by dividing the conditional probability (Eq~\ref{eq:basic}) by a domain-conditional prior probability for the answer choice.

\paragraph{SEQA} \cite{niu-etal-2021-semantic} uses a LM to generate a set of answer candidates. These candidates then ``vote'' for an original answer candidate based on their semantic similarity to each candidate, and the top-voted answer is selected as the final answer. For a fair comparison with the other model, we changed the voting model from SRoBERTa$^{NLI}$ to the origin SRoBERTa that was not further fine-tuned on an NLI dataset.


\begin{table*}[t]
\centering
\small
\begin{tabular}{l|l|l|cccccc}
\toprule
\textbf{Methods} & \multicolumn{2}{c|}{\textbf{LM}} & \textbf{COPA} & \textbf{SCT} & \textbf{SocialIQA} & \textbf{ARC-E} & \textbf{ARC-C} & \textbf{OBQA}\\
\midrule
& \enspace Scoring\quad & Generating \\
\midrule
LM$_{sum}$ & GPT2 & - & 69.0 & 67.6 & 43.1 & 53.5 & 25.4 & 22.4\\
LM$_{avg}$ & GPT2 & - & 68.4 & 71.5 & 45.8 & 47.4 & 28.7 & 30.8 \\
\midrule
Self-talk & GPT2 & GPT2 & 66.2 & 70.4 & 47.5 & - & - & - \\
DCPMI & GPT2 & - & 70.8 & 68.6 & 39.2 & 36.0 & 25.1 & 31.4 \\
SEQA & SRoBERTa & GPT2 & 55.8 & 57.4 & 36.4 & 32.1 & 23.7 & 21.2  \\
SEQA$_{GPT3}$ & SRoBERTa & GPT3 & 66.2 & 64.4 & 40.3 & 54.4 & 34.8 & 22.2  \\
CDG & GPT2 & COMET & 72.2 & 71.5 & 45.4 & - & - & -  \\
ArT & GPT2 & GPT2 & 69.8 & 71.6 & 47.3 & - & - & -  \\
\midrule
\methodnoc & GPT2 & - &70.4 & 73.0 & 46.0 & 55.8 & 28.8 & 32.6\\
\method$_{GPT2}$ & GPT2 & GPT2 & 73.8&  76.1 & 46.1 & 54.4 & 30.8 & 30.2\\
\method$_{GPT3}$ & GPT2 & GPT3 & \textbf{78.2} & \textbf{83.2} & \textbf{48.5} & \textbf{63.2} & \textbf{36.5} & \textbf{35.2}\\
\midrule
\end{tabular}
\centering
\caption{Accuracy (\%) of the scoring various methods on the dev sets. All scoring methods are based on GPT-2$_{xlarge}$. \method$_{GPT2}$ and \method$_{GPT3}$ denote \method{} with candidate generation by GPT-2$_{xlarge}$ and GPT-3 respectively. \textbf{Takeaway}: Weighting leads to substantial improvements. When combined with candidate generation, it outperforms all baselines by a large margin.}
\label{tab:mainresults}
\end{table*}

\paragraph{CDG} \cite{bosselut2021dynamic} uses knowledge from COMET \cite{bosselut-etal-2019-comet} to construct a local commonsense knowledge graph for reasoning and inference.

\paragraph{ArT} \cite{wang-zhao-2022-art} consists of two steps: notes taking and reverse thinking. In the notes taking step, the LM generates templated inferences pertaining to key phrases in the context, which are later added as additional knowledge. The reverse thinking step aggregates the scores of different orders of the answer and question (e.g. ``x because y'' vs. ``y therefore x''). 

\subsection{Setup and Hyper-parameters}
\label{sec:exp:setup}

We used GPT-2 via the HuggingFace Transformers library \cite{wolf-etal-2020-transformers} for the scoring part, and GPT-2 XL and GPT-3 \texttt{davinci-003} for the answer space expansion step. In the keyword extraction step (\S\ref{sec:method_grouping}), we included ConceptNet paths with up to $k = 3$ edges. In the weight assigning step 
(\S\ref{sec:method_weight_assign}) we set the coefficient $\lambda$ to 10. 

In the answer space expansion step (\S\ref{sec:method_generation}), we generated $n_{A^*} = 100$ answers from GPT-2 and $n_{A^*} = 50$ answers from GPT-3 for each question. Similarly to SEQA, we used nucleus sampling \cite{holtzman-etal-2021-surface} with $p=0.9$ and set a maximum length of 15 tokens for both LMs. We set the sentence similarity threshold to $s_{sim} = 0.5$ for GPT2 x-large and $s_{sim} = 0.6$ for GPT-3.

Hyper-parameter values were selected based on preliminary experiments on the training sets and were not tuned on the dev sets.

\section{Results}
\label{sec:results}

\subsection{Main Results}
\label{sec:main results}

The performance of the various scoring methods on the 5 benchmarks are presented in Table~\ref{tab:mainresults}. For fair comparison with the baselines, the table shows the performance when GPT2$_{xlarge}$ is used. We report the accuracy on the dev set. \methodnoc{} stands for Commonsense-Augmented Scoring, i.e. it excludes the candidate generation. 



The performance of \methodnoc{} shows that weighting leads to substantial improvements upon the simpler baselines. \methodnoc{} also stands out in the competition with DCPMI, which can also be regarded as a special weight-scoring method.

When combined with candidate generation, \method{} outperforms nearly all baselines, except for the SocialIQA dataset, on which ArT and Self-talk perform better. Notably, both baselines rely on human-designed prompts to generate additional information, which might give them an advantage. 

The gap in performance from SEQA, which also expands the answer space by generating candidate answers, further demonstrates the effectiveness of dynamic weighting. 


\subsection{Effect of the Scoring LM Size}
\label{sec:different_size}

Table~\ref{tab:model_size} shows the performance of \methodnoc{}, \method{} and the simple baselines when using different sizes of GPT-2 models in the scoring part. 


\paragraph{Bigger is better.} Across the various methods, bigger LMs perform better than smaller LMs. 

\paragraph{Smaller LMs gain more from candidate generation.} While all LMs benefit from weighting and candidate generation, smaller LMs gain bigger improvements. For example, candidate generation with GPT-3 adds 13.4 points on COPA to a GPT2$_S$ \methodnoc{} scorer, but only 8.2 points for GPT2$_{XL}$. We hypothesize that the model performance is more sensitive to the LM quality when a single sentence is considered, while expanding the answer space makes even the lower-quality LMs more robust. 

\begin{table}[t]
    \setlength{\tabcolsep}{4pt} 

    \centering
    \scriptsize
    \begin{tabular}{llcccc}
    \toprule
    \textbf{Dataset} & \textbf{Methods} & \textbf{GPT2}$_{S}$ & \textbf{GPT2}$_{M}$ & \textbf{GPT2}$_{L}$ & \textbf{GPT2}$_{XL}$ \\
    \midrule
    \multirow{5}{*}{COPA} & LM$_{sum}$ & 60.0 & 66.6 & 69.2 & 69.0 \\
    & LM$_{avg}$ & 62.6 & 65.4 & 67.0 & 68.4 \\
    & \methodnoc & 62.0 & 67.2 & 69.4 & 70.4 \\
    & \method$_{GPT2}$ & 69.6 & 72.0 & 72.2 & 73.8 \\
    & \method$_{GPT3}$ & \textbf{75.4} & \textbf{76.4} & \textbf{77.4} & \textbf{78.2} \\
    \midrule
    \multirow{5}{*}{SCT} & LM$_{sum}$ & 58.2 & 62.7 & 64.4 & 67.9 \\
    & LM$_{avg}$ & 60.4 & 66.4 & 68.8 & 71.5 \\
    & \methodnoc & 61.9 & 67.5 & 70.9 & 73.0 \\
    & \method$_{GPT2}$ & 74.0 & 75.2 & 75.7 & 76.1 \\
    & \method$_{GPT3}$ & \textbf{76.7} & \textbf{78.6} & \textbf{79.0} & \textbf{83.2} \\
    \midrule
    \multirow{5}{*}{SIQA} & LM$_{sum}$ & 39.7 & 41.4 & 42.0 & 43.1 \\
    & LM$_{avg}$ & 41.8 & 44.1 & 44.9 & 45.8 \\
    & \methodnoc & 42.8 & 44.6 & 45.7 & 46.0 \\
    & \method$_{GPT2}$ & 43.9 & 43.7 & 44.1 & 44.5 \\
    & \method$_{GPT3}$ & \textbf{47.6} & \textbf{48.4} & \textbf{48.5} & \textbf{48.5} \\
    \midrule
     \multirow{5}{*}{ARC-E} & LM$_{sum}$ & 44.2&  48.8 & 50.4 & 53.5\\
    & LM$_{avg}$ & 37.9 & 40.2 & 45.1 & 47.4\\
    & \methodnoc & 46.1 & 49.8 & 53.0 & 55.8\\
    & \method$_{GPT2}$ & 46.5 & 49.6 & 52.0 & 54.4\\
    & \method$_{GPT3}$ & \textbf{54.2} & \textbf{59.1} & \textbf{60.0} & \textbf{63.2}\\
    \midrule
     \multirow{5}{*}{ARC-C} & LM$_{sum}$ & 19.7 &  23.1 &  22.7 & 25.4\\
    & LM$_{avg}$ & 23.4 & 23.7 & 25.4 & 28.7\\
    & \methodnoc & 26.4 & 26.4 & 27.4 & 28.8 \\
    & \method$_{GPT2}$ & 28.1 & 29.4 & 27.8 & 30.8\\
    & \method$_{GPT3}$ & \textbf{33.4} & \textbf{35.3} & \textbf{33.8} & \textbf{36.5} \\
    \midrule
     \multirow{5}{*}{OBQA} & LM$_{sum}$ & 16.2 & 18.2 & 21.8 & 22.4 \\
    & LM$_{avg}$ & 23.0 & 26.8 & 30.0 & 30.8 \\
    & \methodnoc & 25.6 & 28.6 & 31.4 & 32.6 \\
    & \method$_{GPT2}$ & 26.0 & 26.6 & 27.4 & 30.2\\
    & \method$_{GPT3}$ & \textbf{32.2} & \textbf{35.4} & \textbf{37.4} & \textbf{35.2} \\
    \bottomrule
    \end{tabular}
    \caption{Accuracy when using GPT2 models with different sizes for the scoring.   \textbf{Takeaways}: \methodnoc{} consistently outperforms standard LM scoring methods, and is outperformed by \method{}. For \method{}, the best performance is achieved when using large GPT2 models for scoring and more importantly, GPT3 for candidate generation.}
    \label{tab:model_size}
\end{table}
\begin{figure}[t]
  \centering
  \includegraphics[trim={0.85cm 0.2cm 0.8cm 1.5cm},clip,width=0.5\textwidth]{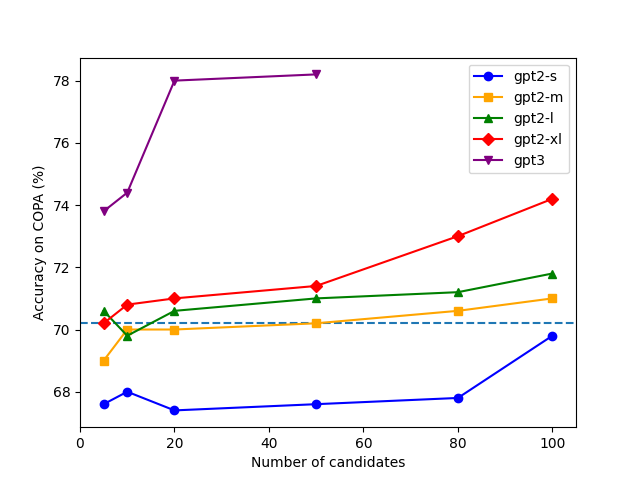}
  \caption{Accuracy curve of \method{} on the COPA dev set, with different numbers of candidates generated from various LMs. The dotted line represents the baseline method LM$_{sum}$ which uses GPT2$_{xlarge}$. \textbf{Takeaways}: Generating more candidates leads to higher accuracy, but larger scoring LMs require fewer candidates.}
  \label{fig:candidate_numbers}
\end{figure}

\subsection{Effect of the No. of Generated Candidates}
\label{sec:candidate_quality}

Figure~\ref{fig:candidate_numbers} shows the effect of the number of generated candidates on the performance, focusing on COPA. We summarize the findings below. 

\paragraph{Generating more candidates leads to higher accuracy.} When generating few ($<20$) candidates, the model's performance is unstable and relatively low. This might happen due to the generated answers being conceptually different from the original candidate answers, in which case they might not meet the mapping thresholds in Section~\ref{sec:method_generation} and be filtered out. This means that  \method{} effectively degenerates to \methodnoc{}. Thus, it's important to generate a large number of candidates. This reassesses the findings in \newcite{niu-etal-2021-semantic}.

\paragraph{Larger models require fewer candidates.} Larger LMs generate higher quality text which is more likely to be fluent, relevant to the context, logically correct, and consistent with commonsense knowledge. Therefore, we can expect fewer candidates to be filtered out. In addition, the generated candidates may be conceptually similar and better phrased than the original choice.


\subsection{Effect of the Weighting Strategy}
\label{sec:static_weight}

Table~\ref{tab:weighting} compares the COPA performance of different weighting strategies. Two baselines, LM$_{sum}$ and LM$_{avg}$, already introduced in Section~\ref{sec:method_basic}, treat all tokens equally, summing or averaging the token-level probabilities. Conversely, the static weighting strategy (\textsc{sw} and \textsc{swc}, with or without candidate generation), assigns a static number (1.5) to each selected key token. Finally, the dynamic weighting strategies  (\methodnoc{} and \method{}) not only distinguish key tokens from unimportant ones but also assign different scores to each key token based on its semantic relevance to the question. 

The results show that while the static weighting strategy outperforms the baseline when no additional candidates are generated (\textsc{sw} vs. LM), these strategies perform similarly when additional candidates are generated (\textsc{swc} vs. LM+c). In both cases, the static weighting strategy underperforms compared to the dynamic strategy. This result confirms that commonsense knowledge can help inform the model about the keywords that are important \emph{for the current question}.

\begin{table}[t]
\centering
\small
\begin{tabular}{lcccc}
\toprule
& GPT2$_{s}$& GPT2$_{m}$& GPT2$_{l}$& GPT2$_{xl}$\\
\midrule
LM$_{sum}$ & 60.0 & 66.6 & 69.2 & 69.0\\
\quad + \textsc{sw} & 61.2 & 66.6 & 70.0 & 69.6\\
\quad + \methodnoc{} & 62.0 & 67.2 & 69.4 & 70.4\\
\hdashline
\quad + \textsc{c} & 69.2 & 71.8 & 70.4 & 72.4 \\
\quad + \textsc{swc} & 69.2 & 71.2 & \textbf{72.4} & 72.2\\
\quad + \method{} & \textbf{69.6} & \textbf{72.0} & 72.2 & \textbf{73.8}\\
\midrule
\end{tabular}
\caption{Accuracy on the COPA dev set when using different weight-assigning methods. The methods below the dotted line expand the answer space by generating additional answer candidates. \textbf{Takeaway}: keyword selection improves the performance, especially when it is informed by commonsense knowledge.}
\label{tab:weighting}
\end{table}

\section{Qualitative Analysis}
\label{sec:analysis}
We focus on \method{} and look at the individual token scores and corresponding ConceptNet paths to better understand the model decision-making process. 



\begin{figure}[t]
  \centering\includegraphics[width=0.48\textwidth,frame]{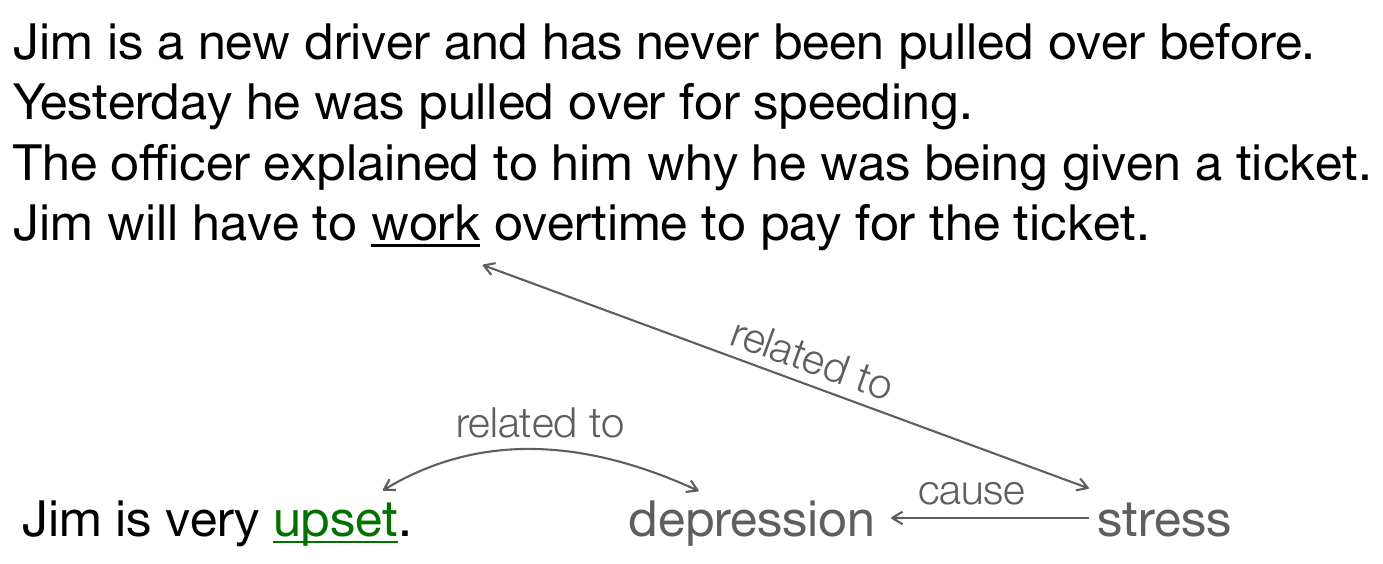}
  \caption{An SCT example, along with the correct answer predicted by \method{}, and an example ConceptNet path that increased the weight of the important word \textit{upset}.}
  \label{fig:success_case}
\end{figure}

Figure~\ref{fig:success_case} shows an example from SCT where \method{} predicted the correct answer. The word ``upset'' in the correct answer choice was assigned a high weight by \method{} thanks to ConceptNet paths such as $\texttt{upset} \xleftrightarrow[]{\text{related to}} \texttt{depression} \xleftarrow[]{\text{causes}} \texttt{stress} \xleftrightarrow[]{\text{related to}} \texttt{work}$. 



\begin{figure}[t]
  \centering
    \includegraphics[width=0.48\textwidth,frame]{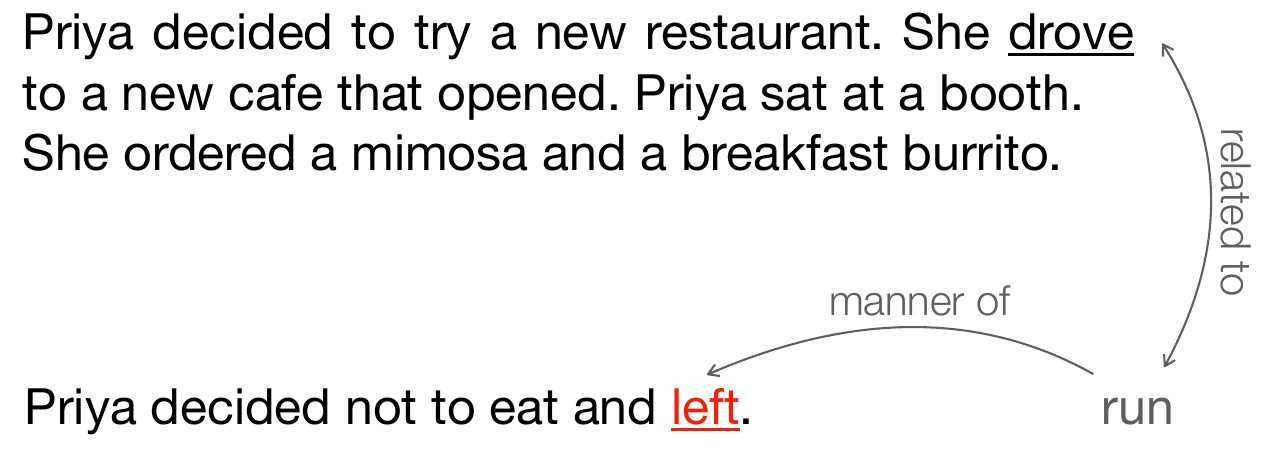}
  \caption{A incorrectly-predicted SCT example, along with the incorrect answer predicted by \method{}, and an example ConceptNet path that increased the weight of the word \textit{left}.}
  \label{fig:failure_case}
\end{figure}

Conversely, in Figure~\ref{fig:failure_case}, \method{} predicted the incorrect answer choice for another SCT example. The model focused on the word ``left'' due to its semantic relation to the word  ``drove'', failing to understand that Priya drove \emph{to} and not \emph{away from} the restaurant. 



\section{Conclusion}
\label{sec:conclusion}

We presented \method{}, a novel LM-based plausibility score for zero-shot MCQA tasks. \method{} uses a commonsense KB to assign importance weights to words in the input. The weighting strategy outperforms basic LM scoring methods. When combined with generating additional answer candidates, \method{} outperforms the baselines on 5 popular MCQA benchmarks. We further showed that the two approaches are complementary and are especially beneficial when using smaller LMs. In the future, we plan to explore a more selective approach for knowledge retrieval from the KB, and adapt \method{} for additional NLP tasks.

\section*{Limitations}
\label{sec:limitations}
\paragraph{Computational complexity.}
\method{} is more computationally expensive than using a basic LM score, as it involves finding relevant paths from an external knowledge base and then estimating their likelihood with a LM, in order to gauge the importance of keywords. 

\paragraph{Concept coverage.} The weight assignment strategy in \method{} is based on ConceptNet. The knowledge in KBs such as ConceptNet is not contextualized, which means that some facts pertaining to concepts in the instance might not be relevant to the specific context. In addition, it has limited coverage. COMET \cite{Hwang2021COMETATOMIC2O} has been used in prior work \cite{majumder-etal-2020-like, chakrabarty-etal-2020-r, Ravi_2023_WACV} to overcome this limitation. However, finding relevant paths using COMET requires an iterative multi-hop reasoning approach \cite{arabshahi-etal-2021-conversational} which is  more complex, and more computationally-intensive. We aim to explore efficient ways to achieve this in future work.


\paragraph{Answer format.} Since our method assigns a weight for each word in the input, it is only applicable for MCQA tasks in which the answer is a sentence. The weighting would be trivial for tasks with single word answers such as  CommonsenseQA \cite{talmor-etal-2019-commonsenseqa} and BoolQ \cite{clark-etal-2019-boolq}. 

\paragraph{Performance limit.} 
Our model demonstrates a significant performance improvement over other zero-shot baselines across a majority of datasets. However, it is worth noting that the state-of-the-art performance on the datasets in this paper is achieved with more supervision (i.e. supervised or few-shot models). 



\section*{Ethics Statement}
\label{sec:ethics}
\paragraph{Data.} All the datasets and knowledge bases used in this work are publicly available. We used ConceptNet as a source of commonsense knowledge. Since ConceptNet was crowdsourced, some of the knowledge may contain societal biases or prejudices held by the annotators \cite{mehrabi-etal-2021-lawyers}. 

\paragraph{Models.} The GPT-2 models are publicly  accessible via HuggingFace, while GPT-3 is a closed model behind an API. All language models may generate offensive statements if prompted with specific inputs, however, our model only generates text internally while the end output is a choice between human-written answer candidates.

\section*{Acknowledgements}
\label{sec:acknowledge}
This work was funded, in part, by the Vector Institute for AI, Canada CIFAR AI Chairs program, an NSERC discovery grant, and a research gift from AI2.

\bibliography{anthology,custom}

\begin{thebibliography}{53}
\expandafter\ifx\csname natexlab\endcsname\relax\def\natexlab#1{#1}\fi

\bibitem[{Abdou et~al.(2020)Abdou, Ravishankar, Barrett, Belinkov, Elliott, and
  S{\o}gaard}]{abdou-etal-2020-sensitivity}
Mostafa Abdou, Vinit Ravishankar, Maria Barrett, Yonatan Belinkov, Desmond
  Elliott, and Anders S{\o}gaard. 2020.
\newblock \href {https://doi.org/10.18653/v1/2020.acl-main.679} {The
  sensitivity of language models and humans to {W}inograd schema
  perturbations}.
\newblock In \emph{Proceedings of the 58th Annual Meeting of the Association
  for Computational Linguistics}, pages 7590--7604, Online. Association for
  Computational Linguistics.

\bibitem[{Arabshahi et~al.(2021)Arabshahi, Lee, Bosselut, Choi, and
  Mitchell}]{arabshahi-etal-2021-conversational}
Forough Arabshahi, Jennifer Lee, Antoine Bosselut, Yejin Choi, and Tom
  Mitchell. 2021.
\newblock \href {https://doi.org/10.18653/v1/2021.emnlp-main.588}
  {Conversational multi-hop reasoning with neural commonsense knowledge and
  symbolic logic rules}.
\newblock In \emph{Proceedings of the 2021 Conference on Empirical Methods in
  Natural Language Processing}, pages 7404--7418, Online and Punta Cana,
  Dominican Republic. Association for Computational Linguistics.

\bibitem[{Bauer et~al.(2018)Bauer, Wang, and
  Bansal}]{bauer-etal-2018-commonsense}
Lisa Bauer, Yicheng Wang, and Mohit Bansal. 2018.
\newblock \href {https://doi.org/10.18653/v1/D18-1454} {Commonsense for
  generative multi-hop question answering tasks}.
\newblock In \emph{Proceedings of the 2018 Conference on Empirical Methods in
  Natural Language Processing}, pages 4220--4230, Brussels, Belgium.
  Association for Computational Linguistics.

\bibitem[{Bosselut et~al.(2021)Bosselut, Le~Bras, and
  Choi}]{bosselut2021dynamic}
Antoine Bosselut, Ronan Le~Bras, and Yejin Choi. 2021.
\newblock Dynamic neuro-symbolic knowledge graph construction for zero-shot
  commonsense question answering.
\newblock In \emph{Proceedings of the AAAI conference on Artificial
  Intelligence}, volume~35, pages 4923--4931.

\bibitem[{Bosselut et~al.(2019)Bosselut, Rashkin, Sap, Malaviya, Celikyilmaz,
  and Choi}]{bosselut-etal-2019-comet}
Antoine Bosselut, Hannah Rashkin, Maarten Sap, Chaitanya Malaviya, Asli
  Celikyilmaz, and Yejin Choi. 2019.
\newblock \href {https://doi.org/10.18653/v1/P19-1470} {{COMET}: Commonsense
  transformers for automatic knowledge graph construction}.
\newblock In \emph{Proceedings of the 57th Annual Meeting of the Association
  for Computational Linguistics}, pages 4762--4779, Florence, Italy.
  Association for Computational Linguistics.

\bibitem[{Brown et~al.(2020{\natexlab{a}})Brown, Mann, Ryder, Subbiah, Kaplan,
  Dhariwal, Neelakantan, Shyam, Sastry, Askell, Agarwal, Herbert-Voss, Krueger,
  Henighan, Child, Ramesh, Ziegler, Wu, Winter, Hesse, Chen, Sigler, Litwin,
  Gray, Chess, Clark, Berner, McCandlish, Radford, Sutskever, and
  Amodei}]{gpt3}
Tom Brown, Benjamin Mann, Nick Ryder, Melanie Subbiah, Jared~D Kaplan, Prafulla
  Dhariwal, Arvind Neelakantan, Pranav Shyam, Girish Sastry, Amanda Askell,
  Sandhini Agarwal, Ariel Herbert-Voss, Gretchen Krueger, Tom Henighan, Rewon
  Child, Aditya Ramesh, Daniel Ziegler, Jeffrey Wu, Clemens Winter, Chris
  Hesse, Mark Chen, Eric Sigler, Mateusz Litwin, Scott Gray, Benjamin Chess,
  Jack Clark, Christopher Berner, Sam McCandlish, Alec Radford, Ilya Sutskever,
  and Dario Amodei. 2020{\natexlab{a}}.
\newblock Language models are few-shot learners.
\newblock In \emph{Advances in Neural Information Processing Systems
  (NeurIPS)}, volume~33, pages 1877--1901.

\bibitem[{Brown et~al.(2020{\natexlab{b}})Brown, Mann, Ryder, Subbiah, Kaplan,
  Dhariwal, Neelakantan, Shyam, Sastry, Askell et~al.}]{brown2020language}
Tom Brown, Benjamin Mann, Nick Ryder, Melanie Subbiah, Jared~D Kaplan, Prafulla
  Dhariwal, Arvind Neelakantan, Pranav Shyam, Girish Sastry, Amanda Askell,
  et~al. 2020{\natexlab{b}}.
\newblock Language models are few-shot learners.
\newblock \emph{Advances in neural information processing systems},
  33:1877--1901.

\bibitem[{Campos et~al.(2018)Campos, Mangaravite, Pasquali, Jorge, Nunes, and
  Jatowt}]{campos2018yake}
Ricardo Campos, V{\'\i}tor Mangaravite, Arian Pasquali, Al{\'\i}pio~M{\'a}rio
  Jorge, C{\'e}lia Nunes, and Adam Jatowt. 2018.
\newblock Yake! collection-independent automatic keyword extractor.
\newblock In \emph{Advances in Information Retrieval: 40th European Conference
  on IR Research, ECIR 2018, Grenoble, France, March 26-29, 2018, Proceedings
  40}, pages 806--810. Springer.

\bibitem[{Chakrabarty et~al.(2022)Chakrabarty, Choi, and
  Shwartz}]{chakrabarty-etal-2022-rocket}
Tuhin Chakrabarty, Yejin Choi, and Vered Shwartz. 2022.
\newblock \href {https://doi.org/10.1162/tacl_a_00478} {It{'}s not rocket
  science: Interpreting figurative language in narratives}.
\newblock \emph{Transactions of the Association for Computational Linguistics},
  10:589--606.

\bibitem[{Chakrabarty et~al.(2020)Chakrabarty, Ghosh, Muresan, and
  Peng}]{chakrabarty-etal-2020-r}
Tuhin Chakrabarty, Debanjan Ghosh, Smaranda Muresan, and Nanyun Peng. 2020.
\newblock \href {https://doi.org/10.18653/v1/2020.acl-main.711} {{R}{\^{}}3:
  Reverse, retrieve, and rank for sarcasm generation with commonsense
  knowledge}.
\newblock In \emph{Proceedings of the 58th Annual Meeting of the Association
  for Computational Linguistics}, pages 7976--7986, Online. Association for
  Computational Linguistics.

\bibitem[{Chen et~al.(2020)Chen, Ji, Chen, and
  Zhang}]{chen-etal-2020-improving}
Qianglong Chen, Feng Ji, Haiqing Chen, and Yin Zhang. 2020.
\newblock \href {https://doi.org/10.18653/v1/2020.coling-main.232} {Improving
  commonsense question answering by graph-based iterative retrieval over
  multiple knowledge sources}.
\newblock In \emph{Proceedings of the 28th International Conference on
  Computational Linguistics}, pages 2583--2594, Barcelona, Spain (Online).
  International Committee on Computational Linguistics.

\bibitem[{Clark et~al.(2019)Clark, Lee, Chang, Kwiatkowski, Collins, and
  Toutanova}]{clark-etal-2019-boolq}
Christopher Clark, Kenton Lee, Ming-Wei Chang, Tom Kwiatkowski, Michael
  Collins, and Kristina Toutanova. 2019.
\newblock \href {https://doi.org/10.18653/v1/N19-1300} {{B}ool{Q}: Exploring
  the surprising difficulty of natural yes/no questions}.
\newblock In \emph{Proceedings of the 2019 Conference of the North {A}merican
  Chapter of the Association for Computational Linguistics: Human Language
  Technologies, Volume 1 (Long and Short Papers)}, pages 2924--2936,
  Minneapolis, Minnesota. Association for Computational Linguistics.

\bibitem[{Clark et~al.(2018)Clark, Cowhey, Etzioni, Khot, Sabharwal, Schoenick,
  and Tafjord}]{clark2018think}
Peter Clark, Isaac Cowhey, Oren Etzioni, Tushar Khot, Ashish Sabharwal, Carissa
  Schoenick, and Oyvind Tafjord. 2018.
\newblock Think you have solved question answering? try arc, the ai2 reasoning
  challenge.
\newblock \emph{arXiv preprint arXiv:1803.05457}.

\bibitem[{Creswell et~al.(2022)Creswell, Shanahan, and
  Higgins}]{creswell2022selection}
Antonia Creswell, Murray Shanahan, and Irina Higgins. 2022.
\newblock Selection-inference: Exploiting large language models for
  interpretable logical reasoning.
\newblock \emph{arXiv preprint arXiv:2205.09712}.

\bibitem[{Davison et~al.(2019)Davison, Feldman, and
  Rush}]{davison-etal-2019-commonsense}
Joe Davison, Joshua Feldman, and Alexander Rush. 2019.
\newblock \href {https://doi.org/10.18653/v1/D19-1109} {Commonsense knowledge
  mining from pretrained models}.
\newblock In \emph{Proceedings of the 2019 Conference on Empirical Methods in
  Natural Language Processing and the 9th International Joint Conference on
  Natural Language Processing (EMNLP-IJCNLP)}, pages 1173--1178, Hong Kong,
  China. Association for Computational Linguistics.

\bibitem[{Fang et~al.(2022)Fang, Wang, Xu, Xu, Sun, Zhu, and
  Zeng}]{fang-etal-2022-leveraging}
Yuwei Fang, Shuohang Wang, Yichong Xu, Ruochen Xu, Siqi Sun, Chenguang Zhu, and
  Michael Zeng. 2022.
\newblock \href {https://doi.org/10.18653/v1/2022.findings-acl.255} {Leveraging
  knowledge in multilingual commonsense reasoning}.
\newblock In \emph{Findings of the Association for Computational Linguistics:
  ACL 2022}, pages 3237--3246, Dublin, Ireland. Association for Computational
  Linguistics.

\bibitem[{Gordon et~al.(2012)Gordon, Kozareva, and
  Roemmele}]{gordon-etal-2012-semeval}
Andrew Gordon, Zornitsa Kozareva, and Melissa Roemmele. 2012.
\newblock \href {https://aclanthology.org/S12-1052} {{S}em{E}val-2012 task 7:
  Choice of plausible alternatives: An evaluation of commonsense causal
  reasoning}.
\newblock In \emph{*{SEM} 2012: The First Joint Conference on Lexical and
  Computational Semantics {--} Volume 1: Proceedings of the main conference and
  the shared task, and Volume 2: Proceedings of the Sixth International
  Workshop on Semantic Evaluation ({S}em{E}val 2012)}, pages 394--398,
  Montr{\'e}al, Canada. Association for Computational Linguistics.

\bibitem[{Guan et~al.(2019)Guan, Wang, and Huang}]{Guan_Wang_Huang_2019}
Jian Guan, Yansen Wang, and Minlie Huang. 2019.
\newblock \href {https://doi.org/10.1609/aaai.v33i01.33016473} {Story ending
  generation with incremental encoding and commonsense knowledge}.
\newblock \emph{Proceedings of the AAAI Conference on Artificial Intelligence},
  33(01):6473--6480.

\bibitem[{Holtzman et~al.(2021)Holtzman, West, Shwartz, Choi, and
  Zettlemoyer}]{holtzman-etal-2021-surface}
Ari Holtzman, Peter West, Vered Shwartz, Yejin Choi, and Luke Zettlemoyer.
  2021.
\newblock \href {https://doi.org/10.18653/v1/2021.emnlp-main.564} {Surface form
  competition: Why the highest probability answer isn{'}t always right}.
\newblock In \emph{Proceedings of the 2021 Conference on Empirical Methods in
  Natural Language Processing}, pages 7038--7051, Online and Punta Cana,
  Dominican Republic. Association for Computational Linguistics.

\bibitem[{Huang et~al.(2021)Huang, He, and Liu}]{huang-etal-2021-improving}
Canming Huang, Weinan He, and Yongmei Liu. 2021.
\newblock \href {https://doi.org/10.18653/v1/2021.findings-emnlp.420}
  {Improving unsupervised commonsense reasoning using knowledge-enabled natural
  language inference}.
\newblock In \emph{Findings of the Association for Computational Linguistics:
  EMNLP 2021}, pages 4875--4885, Punta Cana, Dominican Republic. Association
  for Computational Linguistics.

\bibitem[{Hwang et~al.(2021)Hwang, Bhagavatula, {Le Bras}, Da, Sakaguchi,
  Bosselut, and Choi}]{Hwang2021COMETATOMIC2O}
Jena~D. Hwang, Chandra Bhagavatula, Ronan {Le Bras}, Jeff Da, Keisuke
  Sakaguchi, Antoine Bosselut, and Yejin Choi. 2021.
\newblock Comet-atomic 2020: On symbolic and neural commonsense knowledge
  graphs.
\newblock In \emph{AAAI}.

\bibitem[{Khot et~al.(2020)Khot, Clark, Guerquin, Jansen, and
  Sabharwal}]{khot2020qasc}
Tushar Khot, Peter Clark, Michal Guerquin, Peter Jansen, and Ashish Sabharwal.
  2020.
\newblock Qasc: A dataset for question answering via sentence composition.
\newblock In \emph{Proceedings of the AAAI Conference on Artificial
  Intelligence}, volume~34, pages 8082--8090.

\bibitem[{Kim et~al.(2022)Kim, Joo, Chae, Kim, Hwang, and
  Yeo}]{kim-etal-2022-mind}
Seungone Kim, Se~June Joo, Hyungjoo Chae, Chaehyeong Kim, Seung-won Hwang, and
  Jinyoung Yeo. 2022.
\newblock \href {https://aclanthology.org/2022.coling-1.548} {Mind the gap!
  injecting commonsense knowledge for abstractive dialogue summarization}.
\newblock In \emph{Proceedings of the 29th International Conference on
  Computational Linguistics}, pages 6285--6300, Gyeongju, Republic of Korea.
  International Committee on Computational Linguistics.

\bibitem[{Klein and Nabi(2021)}]{klein-nabi-2021-towards}
Tassilo Klein and Moin Nabi. 2021.
\newblock \href {https://doi.org/10.18653/v1/2021.emnlp-main.688} {Towards
  zero-shot commonsense reasoning with self-supervised refinement of language
  models}.
\newblock In \emph{Proceedings of the 2021 Conference on Empirical Methods in
  Natural Language Processing}, pages 8737--8743, Online and Punta Cana,
  Dominican Republic. Association for Computational Linguistics.

\bibitem[{Lin et~al.(2019)Lin, Chen, Chen, and Ren}]{lin-etal-2019-kagnet}
Bill~Yuchen Lin, Xinyue Chen, Jamin Chen, and Xiang Ren. 2019.
\newblock \href {https://doi.org/10.18653/v1/D19-1282} {{K}ag{N}et:
  Knowledge-aware graph networks for commonsense reasoning}.
\newblock In \emph{Proceedings of the 2019 Conference on Empirical Methods in
  Natural Language Processing and the 9th International Joint Conference on
  Natural Language Processing (EMNLP-IJCNLP)}, pages 2829--2839, Hong Kong,
  China. Association for Computational Linguistics.

\bibitem[{Liu et~al.(2022)Liu, Liu, Lu, Welleck, West, Le~Bras, Choi, and
  Hajishirzi}]{liu-etal-2022-generated}
Jiacheng Liu, Alisa Liu, Ximing Lu, Sean Welleck, Peter West, Ronan Le~Bras,
  Yejin Choi, and Hannaneh Hajishirzi. 2022.
\newblock \href {https://doi.org/10.18653/v1/2022.acl-long.225} {Generated
  knowledge prompting for commonsense reasoning}.
\newblock In \emph{Proceedings of the 60th Annual Meeting of the Association
  for Computational Linguistics (Volume 1: Long Papers)}, pages 3154--3169,
  Dublin, Ireland. Association for Computational Linguistics.

\bibitem[{Majumder et~al.(2020)Majumder, Jhamtani, Berg-Kirkpatrick, and
  McAuley}]{majumder-etal-2020-like}
Bodhisattwa~Prasad Majumder, Harsh Jhamtani, Taylor Berg-Kirkpatrick, and
  Julian McAuley. 2020.
\newblock \href {https://doi.org/10.18653/v1/2020.emnlp-main.739} {Like hiking?
  you probably enjoy nature: Persona-grounded dialog with commonsense
  expansions}.
\newblock In \emph{Proceedings of the 2020 Conference on Empirical Methods in
  Natural Language Processing (EMNLP)}, pages 9194--9206, Online. Association
  for Computational Linguistics.

\bibitem[{Mao et~al.(2019)Mao, Majumder, McAuley, and
  Cottrell}]{mao-etal-2019-improving}
Huanru~Henry Mao, Bodhisattwa~Prasad Majumder, Julian McAuley, and Garrison
  Cottrell. 2019.
\newblock \href {https://doi.org/10.18653/v1/D19-1615} {Improving neural story
  generation by targeted common sense grounding}.
\newblock In \emph{Proceedings of the 2019 Conference on Empirical Methods in
  Natural Language Processing and the 9th International Joint Conference on
  Natural Language Processing (EMNLP-IJCNLP)}, pages 5988--5993, Hong Kong,
  China. Association for Computational Linguistics.

\bibitem[{Mehrabi et~al.(2021)Mehrabi, Zhou, Morstatter, Pujara, Ren, and
  Galstyan}]{mehrabi-etal-2021-lawyers}
Ninareh Mehrabi, Pei Zhou, Fred Morstatter, Jay Pujara, Xiang Ren, and Aram
  Galstyan. 2021.
\newblock \href {https://doi.org/10.18653/v1/2021.emnlp-main.410} {Lawyers are
  dishonest? quantifying representational harms in commonsense knowledge
  resources}.
\newblock In \emph{Proceedings of the 2021 Conference on Empirical Methods in
  Natural Language Processing}, pages 5016--5033, Online and Punta Cana,
  Dominican Republic. Association for Computational Linguistics.

\bibitem[{Mihaylov et~al.(2018)Mihaylov, Clark, Khot, and
  Sabharwal}]{mihaylov-etal-2018-suit}
Todor Mihaylov, Peter Clark, Tushar Khot, and Ashish Sabharwal. 2018.
\newblock \href {https://doi.org/10.18653/v1/D18-1260} {Can a suit of armor
  conduct electricity? a new dataset for open book question answering}.
\newblock In \emph{Proceedings of the 2018 Conference on Empirical Methods in
  Natural Language Processing}, pages 2381--2391, Brussels, Belgium.
  Association for Computational Linguistics.

\bibitem[{Mostafazadeh et~al.(2016)Mostafazadeh, Chambers, He, Parikh, Batra,
  Vanderwende, Kohli, and Allen}]{mostafazadeh-etal-2016-corpus}
Nasrin Mostafazadeh, Nathanael Chambers, Xiaodong He, Devi Parikh, Dhruv Batra,
  Lucy Vanderwende, Pushmeet Kohli, and James Allen. 2016.
\newblock \href {https://doi.org/10.18653/v1/N16-1098} {A corpus and cloze
  evaluation for deeper understanding of commonsense stories}.
\newblock In \emph{Proceedings of the 2016 Conference of the North {A}merican
  Chapter of the Association for Computational Linguistics: Human Language
  Technologies}, pages 839--849, San Diego, California. Association for
  Computational Linguistics.

\bibitem[{Niu et~al.(2021)Niu, Huang, Liang, Chen, Zhu, and
  Huang}]{niu-etal-2021-semantic}
Yilin Niu, Fei Huang, Jiaming Liang, Wenkai Chen, Xiaoyan Zhu, and Minlie
  Huang. 2021.
\newblock \href {https://doi.org/10.18653/v1/2021.acl-long.237} {A
  semantic-based method for unsupervised commonsense question answering}.
\newblock In \emph{Proceedings of the 59th Annual Meeting of the Association
  for Computational Linguistics and the 11th International Joint Conference on
  Natural Language Processing (Volume 1: Long Papers)}, pages 3037--3049,
  Online. Association for Computational Linguistics.

\bibitem[{Petroni et~al.(2019)Petroni, Rockt{\"a}schel, Riedel, Lewis, Bakhtin,
  Wu, and Miller}]{petroni-etal-2019-language}
Fabio Petroni, Tim Rockt{\"a}schel, Sebastian Riedel, Patrick Lewis, Anton
  Bakhtin, Yuxiang Wu, and Alexander Miller. 2019.
\newblock \href {https://doi.org/10.18653/v1/D19-1250} {Language models as
  knowledge bases?}
\newblock In \emph{Proceedings of the 2019 Conference on Empirical Methods in
  Natural Language Processing and the 9th International Joint Conference on
  Natural Language Processing (EMNLP-IJCNLP)}, pages 2463--2473, Hong Kong,
  China. Association for Computational Linguistics.

\bibitem[{Radford et~al.(2019)Radford, Wu, Child, Luan, Amodei, and
  Sutskever}]{gpt2}
Alec Radford, Jeff Wu, Rewon Child, David Luan, Dario Amodei, and Ilya
  Sutskever. 2019.
\newblock Language models are unsupervised multitask learners.

\bibitem[{Ravi et~al.(2023)Ravi, Chinchure, Sigal, Liao, and
  Shwartz}]{Ravi_2023_WACV}
Sahithya Ravi, Aditya Chinchure, Leonid Sigal, Renjie Liao, and Vered Shwartz.
  2023.
\newblock Vlc-bert: Visual question answering with contextualized commonsense
  knowledge.
\newblock In \emph{Proceedings of the IEEE/CVF Winter Conference on
  Applications of Computer Vision (WACV)}, pages 1155--1165.

\bibitem[{Reimers and Gurevych(2019)}]{reimers-gurevych-2019-sentence}
Nils Reimers and Iryna Gurevych. 2019.
\newblock \href {https://doi.org/10.18653/v1/D19-1410} {Sentence-{BERT}:
  Sentence embeddings using {S}iamese {BERT}-networks}.
\newblock In \emph{Proceedings of the 2019 Conference on Empirical Methods in
  Natural Language Processing and the 9th International Joint Conference on
  Natural Language Processing (EMNLP-IJCNLP)}, pages 3982--3992, Hong Kong,
  China. Association for Computational Linguistics.

\bibitem[{Roemmele et~al.(2011)Roemmele, Bejan, and
  Gordon}]{roemmele2011choice}
Melissa Roemmele, Cosmin~Adrian Bejan, and Andrew~S Gordon. 2011.
\newblock Choice of plausible alternatives: An evaluation of commonsense causal
  reasoning.
\newblock In \emph{AAAI spring symposium: logical formalizations of commonsense
  reasoning}, pages 90--95.

\bibitem[{Saha et~al.(2022)Saha, Joty, and Hoi}]{Saha_Joty_Hoi_2022}
Amrita Saha, Shafiq Joty, and Steven~C.H. Hoi. 2022.
\newblock \href {https://doi.org/10.1609/aaai.v36i10.21374} {Weakly supervised
  neuro-symbolic module networks for numerical reasoning over text}.
\newblock \emph{Proceedings of the AAAI Conference on Artificial Intelligence},
  36(10):11238--11247.

\bibitem[{Sap et~al.(2019{\natexlab{a}})Sap, Le~Bras, Allaway, Bhagavatula,
  Lourie, Rashkin, Roof, Smith, and Choi}]{sap2019atomic}
Maarten Sap, Ronan Le~Bras, Emily Allaway, Chandra Bhagavatula, Nicholas
  Lourie, Hannah Rashkin, Brendan Roof, Noah~A Smith, and Yejin Choi.
  2019{\natexlab{a}}.
\newblock Atomic: An atlas of machine commonsense for if-then reasoning.
\newblock In \emph{Proceedings of the AAAI conference on artificial
  intelligence}, volume~33, pages 3027--3035.

\bibitem[{Sap et~al.(2019{\natexlab{b}})Sap, Rashkin, Chen, Le~Bras, and
  Choi}]{sap-etal-2019-social}
Maarten Sap, Hannah Rashkin, Derek Chen, Ronan Le~Bras, and Yejin Choi.
  2019{\natexlab{b}}.
\newblock \href {https://doi.org/10.18653/v1/D19-1454} {Social {IQ}a:
  Commonsense reasoning about social interactions}.
\newblock In \emph{Proceedings of the 2019 Conference on Empirical Methods in
  Natural Language Processing and the 9th International Joint Conference on
  Natural Language Processing (EMNLP-IJCNLP)}, pages 4463--4473, Hong Kong,
  China. Association for Computational Linguistics.

\bibitem[{Shwartz et~al.(2020)Shwartz, West, Le~Bras, Bhagavatula, and
  Choi}]{shwartz-etal-2020-unsupervised}
Vered Shwartz, Peter West, Ronan Le~Bras, Chandra Bhagavatula, and Yejin Choi.
  2020.
\newblock \href {https://doi.org/10.18653/v1/2020.emnlp-main.373} {Unsupervised
  commonsense question answering with self-talk}.
\newblock In \emph{Proceedings of the 2020 Conference on Empirical Methods in
  Natural Language Processing (EMNLP)}, pages 4615--4629, Online. Association
  for Computational Linguistics.

\bibitem[{Speer et~al.(2017)Speer, Chin, and Havasi}]{speer2017conceptnet}
Robyn Speer, Joshua Chin, and Catherine Havasi. 2017.
\newblock Conceptnet 5.5: An open multilingual graph of general knowledge.
\newblock In \emph{Proceedings of the AAAI conference on artificial
  intelligence}, volume~31.

\bibitem[{Talmor et~al.(2019)Talmor, Herzig, Lourie, and
  Berant}]{talmor-etal-2019-commonsenseqa}
Alon Talmor, Jonathan Herzig, Nicholas Lourie, and Jonathan Berant. 2019.
\newblock \href {https://doi.org/10.18653/v1/N19-1421} {{C}ommonsense{QA}: A
  question answering challenge targeting commonsense knowledge}.
\newblock In \emph{Proceedings of the 2019 Conference of the North {A}merican
  Chapter of the Association for Computational Linguistics: Human Language
  Technologies, Volume 1 (Long and Short Papers)}, pages 4149--4158,
  Minneapolis, Minnesota. Association for Computational Linguistics.

\bibitem[{Tamborrino et~al.(2020)Tamborrino, Pellican{\`o}, Pannier, Voitot,
  and Naudin}]{tamborrino-etal-2020-pre}
Alexandre Tamborrino, Nicola Pellican{\`o}, Baptiste Pannier, Pascal Voitot,
  and Louise Naudin. 2020.
\newblock \href {https://doi.org/10.18653/v1/2020.acl-main.357} {Pre-training
  is (almost) all you need: An application to commonsense reasoning}.
\newblock In \emph{Proceedings of the 58th Annual Meeting of the Association
  for Computational Linguistics}, pages 3878--3887, Online. Association for
  Computational Linguistics.

\bibitem[{Touvron et~al.(2023)Touvron, Lavril, Izacard, Martinet, Lachaux,
  Lacroix, Rozière, Goyal, Hambro, Azhar, Rodriguez, Joulin, Grave, and
  Lample}]{touvron2023llama}
Hugo Touvron, Thibaut Lavril, Gautier Izacard, Xavier Martinet, Marie-Anne
  Lachaux, Timothée Lacroix, Baptiste Rozière, Naman Goyal, Eric Hambro,
  Faisal Azhar, Aurelien Rodriguez, Armand Joulin, Edouard Grave, and Guillaume
  Lample. 2023.
\newblock \href {http://arxiv.org/abs/2302.13971} {Llama: Open and efficient
  foundation language models}.

\bibitem[{Trinh and Le(2018)}]{trinh2018simple}
Trieu~H. Trinh and Quoc~V. Le. 2018.
\newblock \href {http://arxiv.org/abs/1806.02847} {A simple method for
  commonsense reasoning}.

\bibitem[{Wang and Zhao(2022)}]{wang-zhao-2022-art}
Jiawei Wang and Hai Zhao. 2022.
\newblock \href {https://aclanthology.org/2022.coling-1.128} {{A}r{T}:
  All-round thinker for unsupervised commonsense question answering}.
\newblock In \emph{Proceedings of the 29th International Conference on
  Computational Linguistics}, pages 1490--1501, Gyeongju, Republic of Korea.
  International Committee on Computational Linguistics.

\bibitem[{Wei et~al.(2022)Wei, Wang, Schuurmans, Bosma, Chi, Le, and
  Zhou}]{wei2022chain}
Jason Wei, Xuezhi Wang, Dale Schuurmans, Maarten Bosma, Ed~Chi, Quoc Le, and
  Denny Zhou. 2022.
\newblock Chain of thought prompting elicits reasoning in large language
  models.
\newblock \emph{arXiv preprint arXiv:2201.11903}.

\bibitem[{Wolf et~al.(2020)Wolf, Debut, Sanh, Chaumond, Delangue, Moi, Cistac,
  Rault, Louf, Funtowicz, Davison, Shleifer, von Platen, Ma, Jernite, Plu, Xu,
  Le~Scao, Gugger, Drame, Lhoest, and Rush}]{wolf-etal-2020-transformers}
Thomas Wolf, Lysandre Debut, Victor Sanh, Julien Chaumond, Clement Delangue,
  Anthony Moi, Pierric Cistac, Tim Rault, Remi Louf, Morgan Funtowicz, Joe
  Davison, Sam Shleifer, Patrick von Platen, Clara Ma, Yacine Jernite, Julien
  Plu, Canwen Xu, Teven Le~Scao, Sylvain Gugger, Mariama Drame, Quentin Lhoest,
  and Alexander Rush. 2020.
\newblock \href {https://doi.org/10.18653/v1/2020.emnlp-demos.6} {Transformers:
  State-of-the-art natural language processing}.
\newblock In \emph{Proceedings of the 2020 Conference on Empirical Methods in
  Natural Language Processing: System Demonstrations}, pages 38--45, Online.
  Association for Computational Linguistics.

\bibitem[{Xia et~al.(2019)Xia, Wu, and Yan}]{xia2019incorporating}
Jiangnan Xia, Chen Wu, and Ming Yan. 2019.
\newblock Incorporating relation knowledge into commonsense reading
  comprehension with multi-task learning.
\newblock In \emph{Proceedings of the 28th ACM International Conference on
  Information and Knowledge Management}, pages 2393--2396.

\bibitem[{Yang et~al.(2018)Yang, Qi, Zhang, Bengio, Cohen, Salakhutdinov, and
  Manning}]{yang-etal-2018-hotpotqa}
Zhilin Yang, Peng Qi, Saizheng Zhang, Yoshua Bengio, William Cohen, Ruslan
  Salakhutdinov, and Christopher~D. Manning. 2018.
\newblock \href {https://doi.org/10.18653/v1/D18-1259} {{H}otpot{QA}: A dataset
  for diverse, explainable multi-hop question answering}.
\newblock In \emph{Proceedings of the 2018 Conference on Empirical Methods in
  Natural Language Processing}, pages 2369--2380, Brussels, Belgium.
  Association for Computational Linguistics.

\bibitem[{Zhang et~al.(2020)Zhang, Williams, Titov, and
  Sennrich}]{zhang-etal-2020-improving}
Biao Zhang, Philip Williams, Ivan Titov, and Rico Sennrich. 2020.
\newblock \href {https://doi.org/10.18653/v1/2020.acl-main.148} {Improving
  massively multilingual neural machine translation and zero-shot translation}.
\newblock In \emph{Proceedings of the 58th Annual Meeting of the Association
  for Computational Linguistics}, pages 1628--1639, Online. Association for
  Computational Linguistics.

\bibitem[{Zhou et~al.(2020)Zhou, Zhang, Cui, and Huang}]{zhou2020evaluating}
Xuhui Zhou, Yue Zhang, Leyang Cui, and Dandan Huang. 2020.
\newblock Evaluating commonsense in pre-trained language models.
\newblock In \emph{Proceedings of the AAAI conference on artificial
  intelligence}, volume~34, pages 9733--9740.

\end{thebibliography}
\bibliographystyle{acl_natbib}

\appendix
\newpage
\section{Question Prompts}
\label{sec:appendix_declarative}

Table~\ref{tab:converted_question} shows the prompts used for each dataset. For tasks with several specific question type such as COPA and SocialIQa, we convert each question type to a natural language proxy following previous work \cite[e.g.][]{shwartz-etal-2020-unsupervised}. For tasks that present an open-ended question, we append the prefix ``The answer is''. Finally, for tasks that are already designed to expect the next word or sentence (such as SCT), we use the instance as is. 

\begin{table}[h]
    \centering
    \small
    \setlength{\tabcolsep}{2pt} 
    \begin{tabular}{lp{5.9cm}}
    \toprule
    \textbf{Dataset} & \textbf{Question} \\ 
    \midrule
    \multirow{2}{*}{COPA} & My body cast a shadow over the grass [\textcolor{red}{because}] \\
    \hhline{~-}
    & The physician misdiagnosed the patient [\textcolor{red}{so}]\\
    \midrule
    \multirow{3}{*}{SCT} & Tyler went to a baseball game. He saw his favorite team! His team played hard. His team won! []\\
    \midrule
    \multirow{2}{*}{SocialIQa} & Tracy didn't go home that evening and resisted Riley's attacks. [\textcolor{red}{Before, Tracy needed to}]\\
    \midrule
    \multirow{4}{*}{ARC} & Which technology was developed most recently? [\textcolor{red}{the answer is}]\\
    \hhline{~-}
    & A green plant absorbs light. A frog eats flies. These are both examples of how organisms []\\
    \midrule
    \multirow{4}{*}{OBQA} & A person can grow cabbage in January with the help of what product? [\textcolor{red}{the answer is}]\\
    \hhline{~-}
    & Gas can fill any container it is given, and liquid []\\
    \bottomrule
    \end{tabular}
    \caption{Question formats used for each dataset. The red words in square brackets are additions to the context, designed specifically for each dataset.}
    \label{tab:converted_question}
\end{table}

\newpage

\section{Relation Templates}
\label{sec:appendix:path_sents}

Table~\ref{tab:converted_relation} displays the templates we used to convert edges with different relation types in ConceptNet to natural language sentences, following \newcite{davison-etal-2019-commonsense}.

\begin{table}[!hb] 
    \scriptsize
    \centering
    \setlength{\tabcolsep}{3pt} 
    \begin{tabular}{ll}
    \toprule
    Relation Type & Template \\ 
    \midrule
    $A \xleftrightarrow[]{\text{related to}} B$ & A is related to B \\
    $A \xrightarrow[]{\text{form of}} B$ & A is a form of B \\
    $A \xrightarrow[]{\text{is a}} B$ & A is a B \\
    $A \xrightarrow[]{\text{part of}} B$ & A is a part of B \\
    $A \xrightarrow[]{\text{has a}} B$ & A has a B \\
    $A \xrightarrow[]{\text{used for}} B$ & A is used for B \\
    $A \xrightarrow[]{\text{not used for}} B$ & A is not used for B \\
    $A \xrightarrow[]{\text{capable of}} B$ & A is capable of B \\
    $A \xrightarrow[]{\text{not capable of}} B$ & A is not capable of B \\
    $A \xrightarrow[]{\text{at location}} B$ & A is a location for B \\
    $A \xrightarrow[]{\text{causes}} B$ & A causes B \\
    $A \xleftarrow[]{\text{has subevent}} B$ & B happens as a subevent of A \\
    $A \xrightarrow[]{\text{has first subevent}} B$ & A begins with B \\
    $A \xrightarrow[]{\text{has last subevent}} B$ & A ends with B \\
    $A \xleftarrow[]{\text{has prerequisite}} B$ & B is a dependency of A \\
    $A \xrightarrow[]{\text{has property}} B$ & A can be described as B \\
    $A \xrightarrow[]{\text{not has property}} B$ & A can not be described as B \\
    $A \xrightarrow[]{\text{motivated by goal}} B$ & Someone does A because they want result B \\
    $A \xrightarrow[]{\text{obstructed by}} B$ & A is a obstacle in the way of B \\
    $A \xrightarrow[]{\text{desires}} B$ & A desires B \\
    $A \xrightarrow[]{\text{not desires}} B$ & A do not desire B \\
    $A \xrightarrow[]{\text{created by}} B$ & A is created by B \\
    $A \xleftrightarrow[]{\text{synonym}} B$ & A is similar to B \\
    $A \xleftrightarrow[]{\text{antonym}} B$ & A is opposite to B \\
    $A \xleftrightarrow[]{\text{distinct from}} B$ & A is distinct from B \\
    $A \xrightarrow[]{\text{derived from}} B$ & A is derived from B \\
    $A \xrightarrow[]{\text{symbol of}} B$ & A is a symbol of B \\
    $A \xrightarrow[]{\text{defined as}} B$ & A is defined as B \\
    $A \xrightarrow[]{\text{manner of}} B$ & A is a specific way to do B \\
    $A \xleftrightarrow[]{\text{located near}} B$ & A is near to B \\
    $A \xrightarrow[]{\text{has context}} B$ & A is a word used in the context of B \\
    $A \xleftrightarrow[]{\text{similar to}} B$ & A is similar to B \\
    $A \xleftrightarrow[]{\text{etymologically related to}} B$ & A have a common origin with B \\
    $A \xrightarrow[]{\text{etymologicallyderivedfrom}} B$ & A is derived from B \\
    $A \xrightarrow[]{\text{causes desire}} B$ & A makes someone want B \\
    $A \xrightarrow[]{\text{made of}} B$ & A is made of B \\
    $A \xleftarrow[]{\text{receives action}} B$ & B can be done to A \\
    \bottomrule
    \end{tabular}
    \caption{Natural language templates for each relation type in ConceptNet.}
    \label{tab:converted_relation}
\end{table}

\end{document}